\pdfoutput=1
\RequirePackage{fix-cm}
\RequirePackage{amsmath} %
\documentclass[twocolumn]{svjour3}           %
\smartqed  %
\usepackage{graphicx}
\usepackage{picinpar}  %
\journalname{K\"unstliche Intelligenz}  %

\usepackage{graphicx}            %
\usepackage{amssymb}             %
\usepackage{amsfonts}            %
\usepackage{enumitem}            %
\usepackage{multirow}            %
\usepackage{siunitx}             %
\usepackage{url}                 %
\usepackage{xspace}              %
\usepackage[T1]{fontenc}         %
\usepackage[hidelinks]{hyperref} %

\graphicspath{{images/}}
\DeclareGraphicsExtensions{.pdf,.png,.jpg,.jpeg}

\newcommand{\degreem}{^{\circ}} %

\newcommand{\figlabel}[1]{\label{fig:#1}}
\newcommand{\tablabel}[1]{\label{tab:#1}}

\newcommand{\figref}[1]{Fig.~\ref{fig:#1}\xspace}
\newcommand{\tabref}[1]{Table~\ref{tab:#1}\xspace}

\newcommand{\nop}{NimbRo\protect\nobreakdash-OP\xspace}
\newcommand{\dop}{DARwIn\protect\nobreakdash-OP\xspace}
\newcommand{\rop}{ROBOTIS OP2\xspace}
\newcommand{\nao}{Nao\xspace}
\newcommand{\cm}{CM730\xspace}
\newcommand{\cmnew}{CM740\xspace}
\newcommand{\itwoc}{I\textsuperscript{2}C\xspace}
\newcommand{\igus}{igus\textsuperscript{\tiny\circledR}\xspace}
\newcommand{\iguhop}{igus\textsuperscript{\tiny\circledR}$\!$ Humanoid Open Platform\xspace}
\newcommand{\iguhopp}{igus\textsuperscript{\tiny\circledR} Humanoid Open Platform\xspace}

\newcommand{\term}[1]{\emph{#1}\xspace}
\newcommand{\degree}{$\degreem$\xspace}

\usepackage{eso-pic}

\AtBeginDocument{\AddToShipoutPictureFG*{\AtTextUpperLeft{\put(0,\LenToUnit{54pt}){\parbox{\textwidth}{\centering\bfseries
German Journal on Artificial Intelligence (KI), volume 30, issue 3, 2016
}}}}}
\begin{document}

\title{The igus Humanoid Open Platform\thanks{This work was supported by grant BE 2556/10 of the German Research Foundation (DFG).}}
\subtitle{A Child-sized 3D Printed Open-Source Robot for Research}

\author{Philipp Allgeuer \and Hafez Farazi \and Grzegorz Ficht \and Michael Schreiber \and Sven Behnke}

\institute{All authors are at:\at
Rheinische Friedrich-Wilhelms-Universit{\"a}t Bonn\\
Friedrich-Ebert-Allee 144, 53113 Bonn\\
\email{pallgeuer@ais.uni-bonn.de}
}

\date{Received: date / Accepted: date}

\maketitle

\begin{abstract}
The use of standard robotic platforms can accelerate research and lower the 
entry barrier for new research groups. There exist many affordable humanoid 
standard platforms in the lower size ranges of up to 60\,cm, but larger humanoid 
robots quickly become less affordable and more difficult to operate, maintain 
and modify. The \iguhopp is a new and affordable, fully open-source humanoid 
platform. At 92\,cm in height, the robot is capable of interacting in an 
environment meant for humans, and is equipped with enough sensors, actuators and 
computing power to support researchers in many fields. The structure of the 
robot is entirely 3D printed, leading to a lightweight and visually appealing 
design. The main features of the platform are described in this article. 
\end{abstract}

\section{Introduction}

\begin{figure}[!t]
\parbox{\linewidth}{\centering
\raisebox{8pt}{\includegraphics[width=0.38\linewidth]{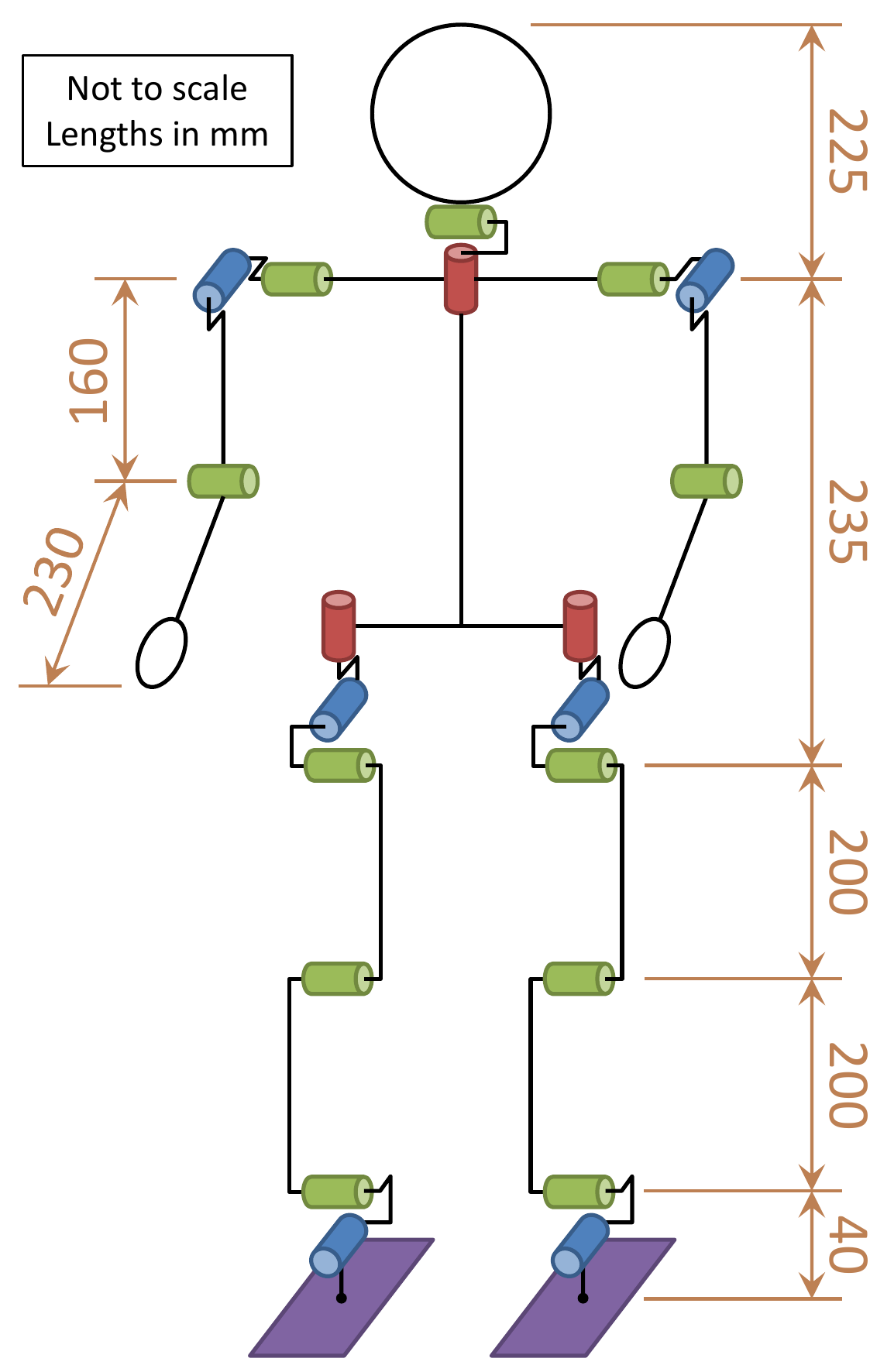}}\hspace{0.06\linewidth}
\includegraphics[width=0.38\linewidth]{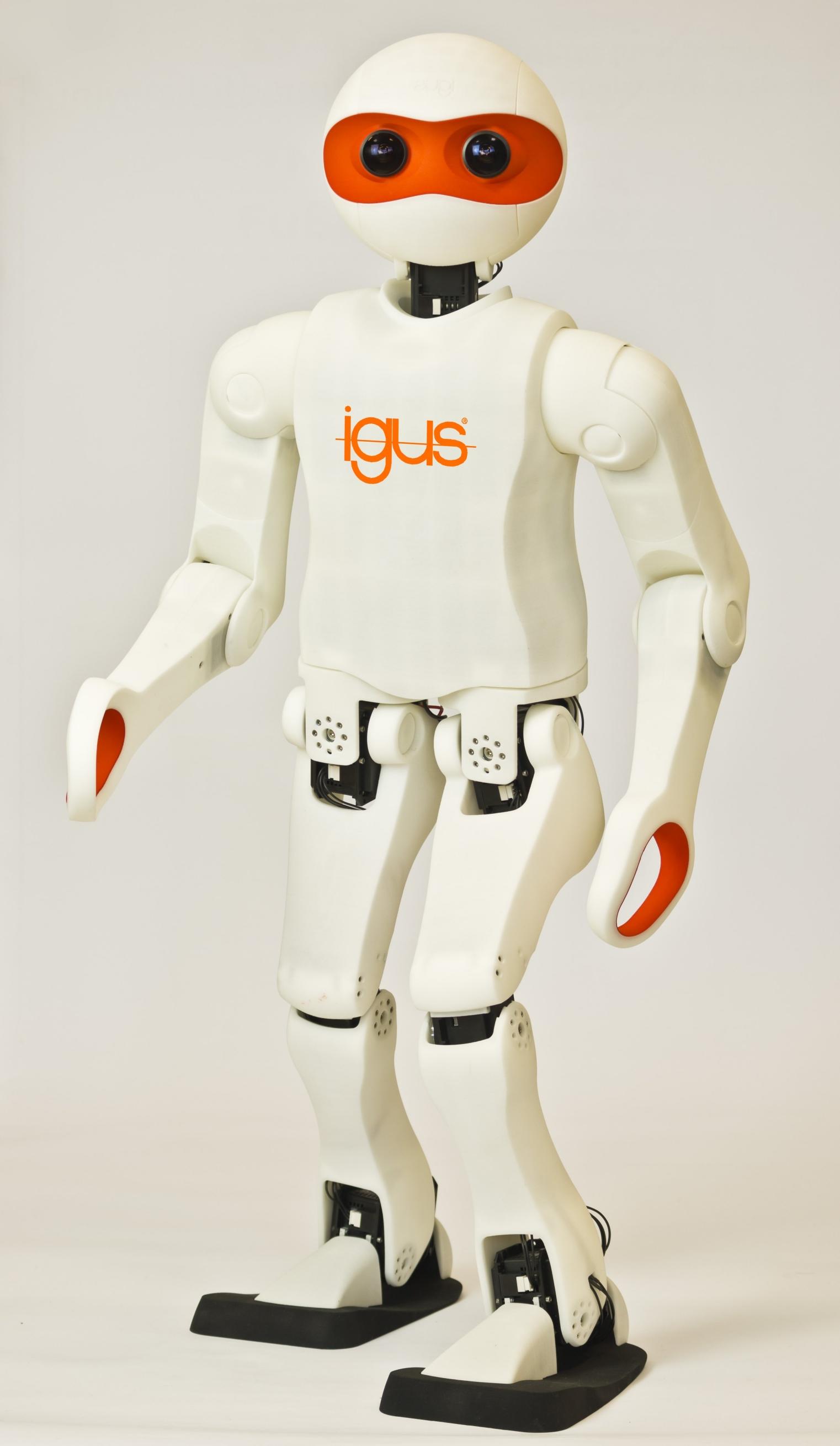}}
\caption{The \iguhop.}
\figlabel{P1_teaser}
\vspace{-3ex}
\end{figure}

The field of humanoid robotics is enjoying increasing popularity, with many 
research groups having developed platforms of all sizes and levels of complexity 
to investigate topics such as bipedal walking, environmental perception, object 
manipulation, and human-machine interaction. The initial effort to start 
humanoid robotics research on a real platform can be high though. Access to a 
standard robot platform can allow for greater focus on research, and facilitates 
greater collaboration and code exchange. The \iguhop, shown in 
\figref{P1_teaser}, is a collaboration between researchers at the University of 
Bonn and \igus GmbH, a leading manufacturer of polymer bearings and energy 
chains. The \iguhop seeks to close the gap between small, albeit affordable, 
standard humanoid platforms, and larger significantly more expensive ones. We 
designed the platform to be as open, modular, maintainable and customisable as 
possible. The use of almost exclusively 3D printed plastic parts for the 
mechanical components of the robot is the result of this mindset, which also 
simplifies the manufacture of the robots. This allows individual parts to be 
easily modified, reprinted and replaced to extend the capabilities of the robot. 
A demonstration video of the \iguhop is available.\footnote{Video: 
\url{https://www.youtube.com/watch?v=RC7ZNXclWWY}}

\section{Related Work}

\begin{table}
\renewcommand{\arraystretch}{1.2}
\caption{\vspace{-0.5ex}\iguhop specifications\vspace{-0.5ex}}
\tablabel{P1_specs}
\centering
\scriptsize
\begin{tabular}{c c c}
\hline
\textbf{Type} & \textbf{Specification} & \textbf{Value}\\
\hline
\multirow{4}{*}{\textbf{General}} & Height \& Weight & \SI{92}{cm}, \SI{6.6}{kg}\\
& Battery & 4-cell LiPo (\SI{14.8}{V}, \SI{3.8}{Ah})\\
& Battery Life & \SI{15}{}--\SI{30}{\minute}\\
& Material & Polyamide 12 (PA12)\\
\hline
\multirow{5}{*}{\textbf{PC}} & Product & Gigabyte Brix GB-BXi7-5500\\
& CPU & Intel i7-5500U, \SI{2.4}{}--\SI{3.0}{GHz}\\
& Memory & \SI{4}{GB} RAM, \SI{120}{GB} SSD\\
& Network & Ethernet, Wi-Fi, Bluetooth\\
& Other & 4$\,\times\,$USB 3.0, HDMI, MiniDP\\
\hline
\multirow{3}{*}{\textbf{\cm}} & Microcontroller & STM32F103RE (Cortex M3)\\
& Memory & \SI{512}{KB} Flash, \SI{64}{KB} SRAM\\
& Other & 3$\,\times\,$Buttons, 7$\,\times\,$LEDs\\
\hline
\multirow{4}{*}{\textbf{Actuators}} & Total & 8$\,\times\,$MX-64, 12$\,\times\,$MX-106\\
& Head & 2$\,\times\,$MX-64\\
& Each Arm & 3$\,\times\,$MX-64\\
& Each Leg & 6$\,\times\,$MX-106\\
\hline
\multirow{6}{*}{\textbf{Sensors}} & Encoders & \SI{4096}{ticks/rev}\\
& Gyroscope & 3-axis (L3G4200D chip)\\
& Accelerometer & 3-axis (LIS331DLH chip)\\
& Magnetometer & 3-axis (HMC5883L chip)\\
& Camera & Logitech C905 (720p)\\
& Camera Lens & Wide-angle lens, 150\degree\!FOV\\
\hline
\end{tabular}
\vspace{-4ex}
\end{table}

A number of standard humanoid robot platforms have been developed over the last 
decade, many of which have seen much success. The most prominent example of this 
is the \nao robot \cite{Gouaillier2009}, developed by Aldebaran Robotics. The 
\nao comes with a rich set of features, such as a variety of available gaits, a 
programming SDK, and human-machine interaction modules. The robot however has a 
limited scope of use as it is only \SI{58}{cm} tall. Also, as a proprietary 
product, there is not much space for own hardware repair and enhancements. 
Another example is the \dop \cite{Ha2011}, and its successor the \rop, 
distributed by Robotis. Both robots are quite similar in design and 
architecture, and stand at \SI{45.5}{cm} tall, half the size of the \iguhop. The 
\dop has the benefit of being an open platform, but its size remains a limiting 
factor for its range of applications.

Other significantly less widely disseminated robots include the Intel Jimmy 
robot, the Poppy robot from the Inria Flowers Laboratory \cite{Lapeyre2014}, and 
the Jinn-Bot from Jinn-Bot Robotics \& Design GmbH in Switzerland. All of these 
robots are at least in part 3D printed, and the first two are open source. The 
Jimmy robot is intended for social interactions and comes with software based on 
the \dop framework. The Poppy robot is intended for non-autonomous use, and 
features a multi-articulated bio-inspired morphology. Jinn-Bot is built from 
over 90 plastic parts and 24 actuators, making for a complicated build, and is 
controlled by a Java application running on a smartphone mounted in its head. 
Larger standard platforms, such as the Asimo \cite{Hirai1998}, HRP 
\cite{Kaneko2009} and Atlas robots, are an order of magnitude more expensive 
and more troublesome to operate and maintain. Such large robots are less robust 
because of their complex hardware structure, and require a gantry in normal use. 
These factors limit the possibility of using such robots by most research 
groups.

\section{Hardware Design}

The \iguhopp was developed with the assistance of a design bureau, to create a 
good overall aesthetic appearance. The main criteria for the design were the 
simplicity of manufacture, assembly, maintenance and customisation. To satisfy 
these criteria, a modular design approach was used. Due to the 3D printed nature 
of the robot, parts can be modified and replaced with great freedom. A summary 
of the hardware specifications of the robot is shown in \tabref{P1_specs}. 

\paragraph{Mechanical Structure}
The white plastic robot exoskeleton is meant not only for outward appearance, 
but also acts as a load-bearing frame. This makes the \iguhop very light in 
comparison to its size. Despite its low weight, the robot is still very durable 
and resistant to deformation and bending. This is achieved through wall 
thickness modulation in the areas more susceptible to damage, as well as through 
strategic distribution of ribs and other strengthening components, which are 
printed as part of the exoskeleton. Due to the versatile nature of 3D printing, 
if a weak spot is identified through practical experience, as indeed happened 
during early testing, the parts can be locally strengthened in the CAD design, 
without significantly impacting the design.

\begin{figure}[!t]
\parbox{\linewidth}{\centering\includegraphics[height=0.78\linewidth,width=0.95\linewidth]{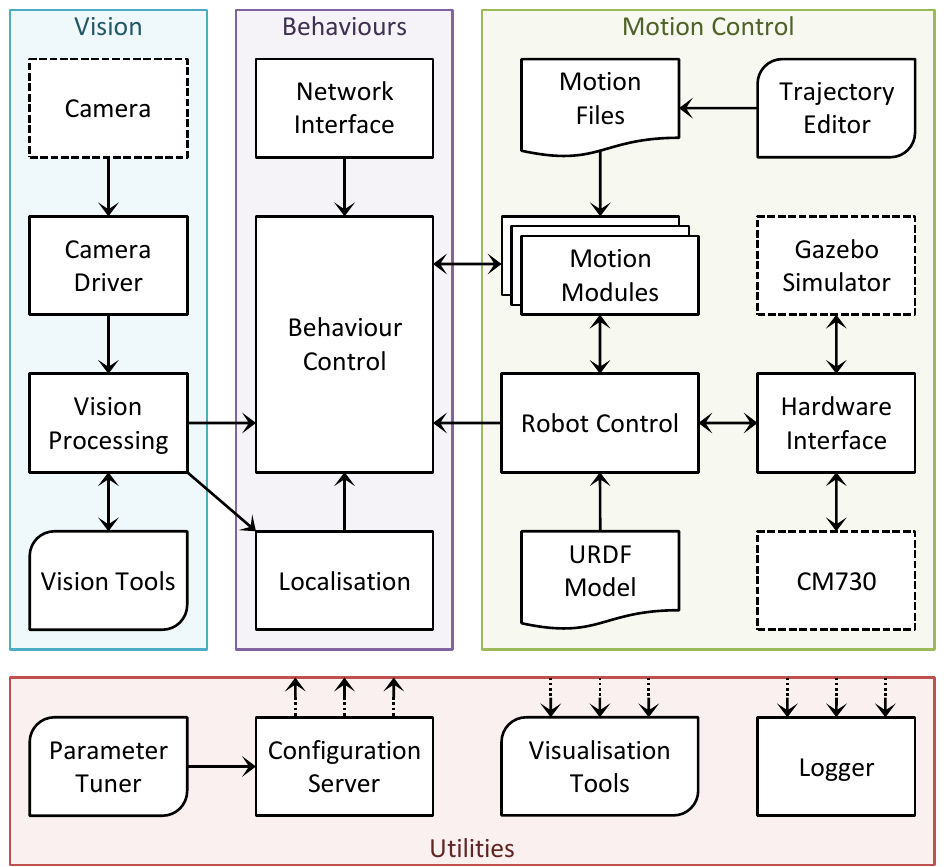}}
\caption{Architecture of the ROS software.}
\figlabel{software_architecture}
\vspace{3ex} %
\end{figure}
\paragraph{Robot Electronics and Sensors}
The electronics of the platform are built around an Intel i7-5500U processor, 
running a full 64-bit Ubuntu OS, and all of the robot control software. DC power 
is provided via a power board, where one or both DC power and a 4-cell Lithium 
Polymer (LiPo) battery can be connected, and the higher voltage of the two is 
used. The PC communicates with a Robotis \cm subcontroller board, whose main 
purpose is to electrically interface the twelve MX-106 and eight MX-64 
actuators, all connected on a single Dynamixel bus.

Due to a number of reliability and performance factors, we redesigned and 
rewrote the firmware of the \cm. This improved bus stability and error 
tolerance, and decreased the time required for the reading out of servo data, 
while still retaining compatibility with the standard Dynamixel protocol. The 
new firmware is compatible with both the \cm and its successor \cmnew. The \cm 
also connects to an interface board that has three buttons, five LEDs and two 
RGB LEDs, and internally incorporates a 3-axis gyroscope and accelerometer. An 
additional 3-axis magnetometer is connected via an \itwoc interface on the 
onboard microcontroller which in total provides the user with a 9-axis IMU.

Further available external connections to the robot include USB, HDMI, Mini 
DisplayPort, Gigabit Ethernet, IEEE 802.11b/g/n Wi-Fi, and Bluetooth 4.0. The 
\iguhop is nominally equipped with a single 720p Logitech C905 camera behind its 
right eye, fitted with a wide-angle lens. A second camera can be optionally 
mounted behind the left eye for stereo vision.

\section{Software}

The ROS middleware was chosen as the basis of the software developed for the 
\iguhop. This fosters modularity, visibility, reusability, and to some degree 
also the platform independence. An overview of the software architecture is 
shown in \figref{software_architecture}. The software was developed with 
humanoid robot soccer in mind, but the platform can be used for virtually any 
other application. This is possible because of the strongly modular way in which 
the software was written, greatly supported by the natural modularity of ROS, 
and the use of plugin schemes.

\paragraph{Vision}
The nominally $640\!\times\!480$ images captured by the camera at \SI{30}{Hz} 
are first converted into the HSV colour space. In our target application of 
soccer, the vision processing tasks include field, ball, goal, field line, 
centre circle and obstacle detection \cite{Farazi2015}, as illustrated in \figref{vision_output}. 
The wide-angle camera used in the platform introduces significant distortion, 
which must be compensated when projecting image coordinates into egocentric 
world coordinates. We undistort the image with a Newton-Raphson approach. This 
method is used to populate lookup tables that allow constant time distortion and 
undistortion at runtime. The effect of undistorting the image is shown in 
\figref{vision_output}. We also compensate for remaining projection errors by 
calibrating the position and orientation of the camera frame using ground 
truth observations and the Nelder-Mead method. This calibration is essential for 
good performance of the projection operations. 

\begin{figure}[!t]
\parbox{\linewidth}{\centering
\includegraphics[width=0.48\linewidth,height=0.32\linewidth]{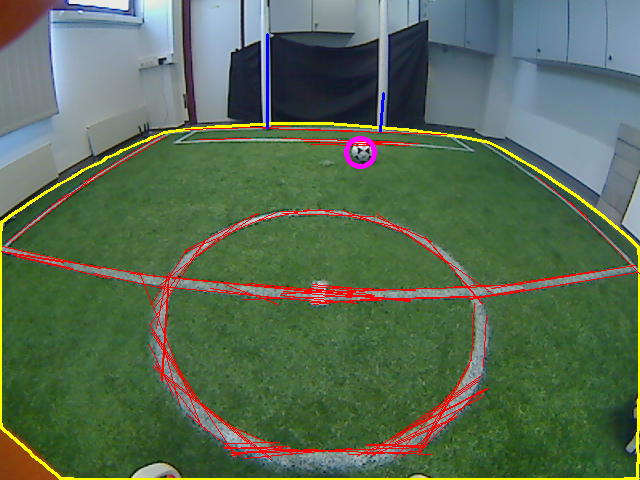}\hspace{0.019\linewidth}
\includegraphics[width=0.48\linewidth,height=0.32\linewidth]{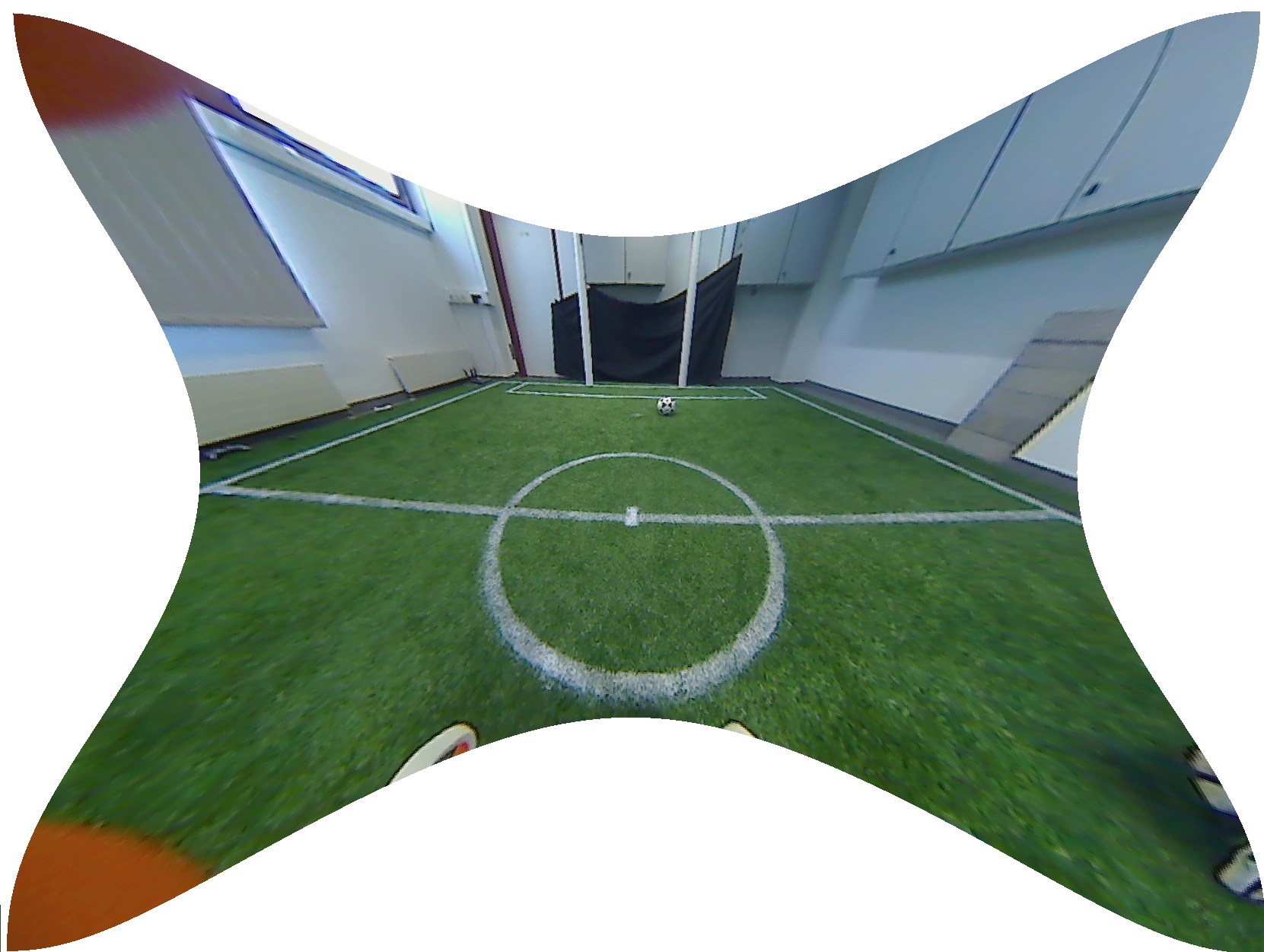}}
\caption{Left: A captured image with ball (pink circle), field line (red lines), 
field boundary (yellow lines), and goal post (blue lines) detections annotated.
Right: The raw captured image with undistortion applied.}
\figlabel{vision_output}
\vspace{-3ex}
\end{figure}
\paragraph{State Estimation}
The 9-axis IMU on the \cm is used to obtain the 3D orientation of the robot 
through the means of a nonlinear passive complementary filter 
\cite{Allgeuer2014}. This filter returns the estimated 3D orientation of the 
robot with the use of a novel way of representing orientations, namely the 
\term{fused angles} representation \cite{Allgeuer2015}.

\paragraph{Actuator Control}
The ability of the actuators to track their set position are influenced by many 
factors, including battery voltage, joint friction, inertia and load. To 
minimise the impact of these factors, we apply feed-forward control to the 
commanded positions \cite{schwarz2013compliant}. This allows the joints to be 
operated in higher ranges of compliance, reduces servo overheating and wear, 
increases battery life, and reduces the problems posed by impacts and 
disturbances. The vector of desired feed-forward output torques is computed from 
the commanded joint positions, velocities and accelerations using the full-body 
inverse dynamics of the robot. The torques are then converted into joint target 
offsets that are added to the setpoints of the position-controlled actuators.

\paragraph{Motions}
Often there is a need for a robot to play a particular pre-designed motion. This 
is the task of the \emph{motion player}, which implements a nonlinear keyframe 
interpolator that connects robot poses, smoothly interpolates joint positions 
and velocities, and modulates the joint efforts and support coefficients. A PID 
feedback scheme is implemented on top of this, where joints can be controlled in 
response to the estimated robot orientation. To create and edit the motions, a 
trajectory editor was developed for the \iguhop. All motions can be edited in a 
user-friendly environment with a 3D preview of the robot poses. We have designed 
numerous motions including kicking, waving, balancing, get-up, and other 
motions, some of which are shown in \figref{P1_getup}.

\begin{figure}[!t]
\parbox{\linewidth}{\centering\includegraphics[width=0.9\linewidth, height=0.44\linewidth]{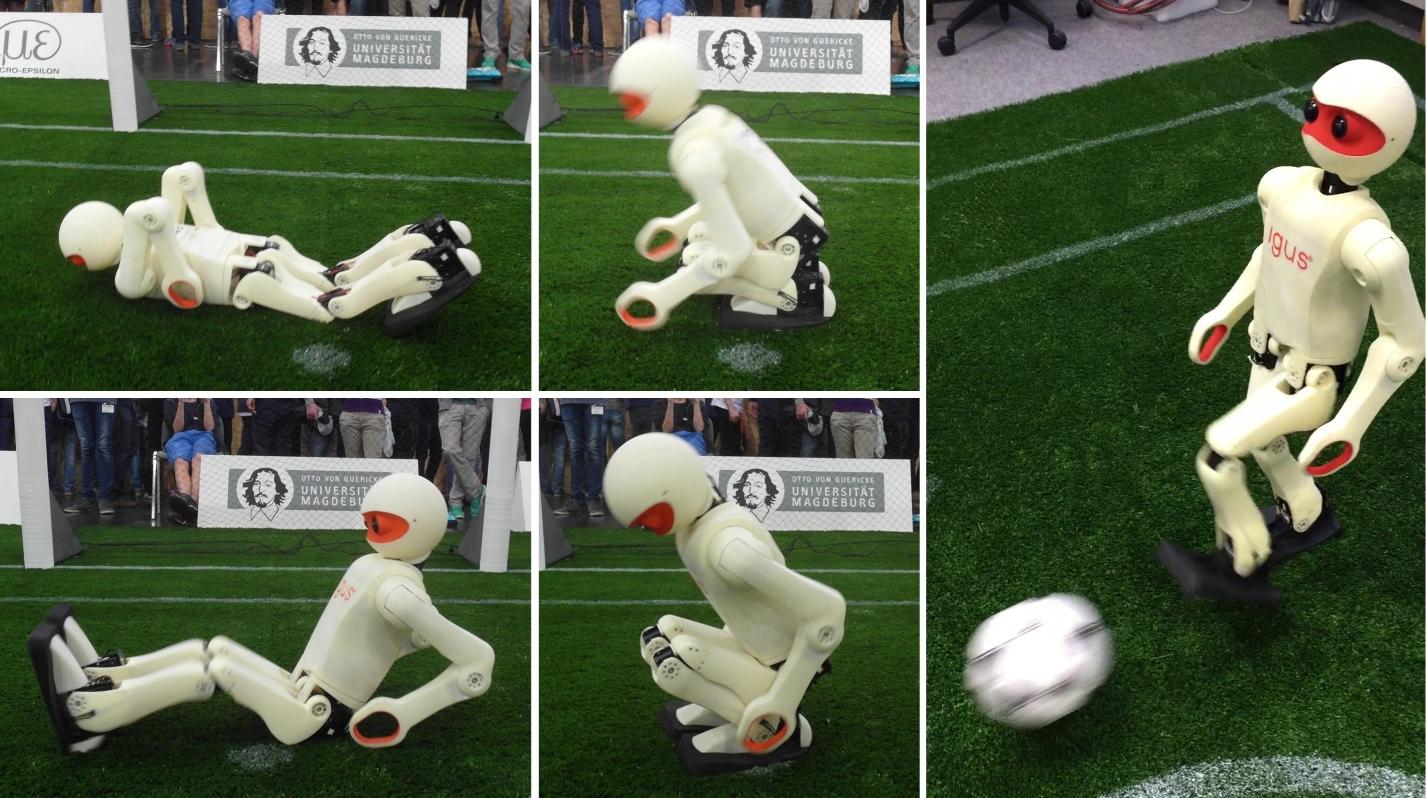}}
\caption{Dynamic get-up motions of the \iguhop, from the prone (top row) and supine
(bottom row) lying positions, and a still image of the dynamic kick motion.}
\figlabel{P1_getup}
\vspace{-3ex}
\end{figure}
\paragraph{Gait Generation}
The gait is formulated in three pose spaces: Joint space, abstract space, and 
inverse space. The \emph{joint space} specifies all of the joint angles, while 
the \emph{inverse space} specifies the Cartesian coordinates and quaternion 
orientations of each of the limb end effectors relative to the trunk link frame. 
The \emph{abstract space}, however is a representation that was specifically 
developed for humanoid robots in the context of walking and balancing 
\cite{Behnke2006}. The walking gait is based on an open loop central pattern 
generated core that is calculated from a gait phase angle that increments at a 
rate proportional to the desired gait frequency. A number of simultaneously 
operating basic feedback mechanisms have been built around the open loop gait 
core to stabilise the walking, illustrated in \figref{basic_feedback}. The 
feedback in each of these mechanisms derives from the fused pitch and fused roll 
state estimates, and adds corrective action components to the central pattern 
generated waveforms in both the abstract and inverse spaces \cite{Allgeuer2015}.

\begin{figure}[!t]
\parbox{\linewidth}{\centering\includegraphics[width=0.97\linewidth]{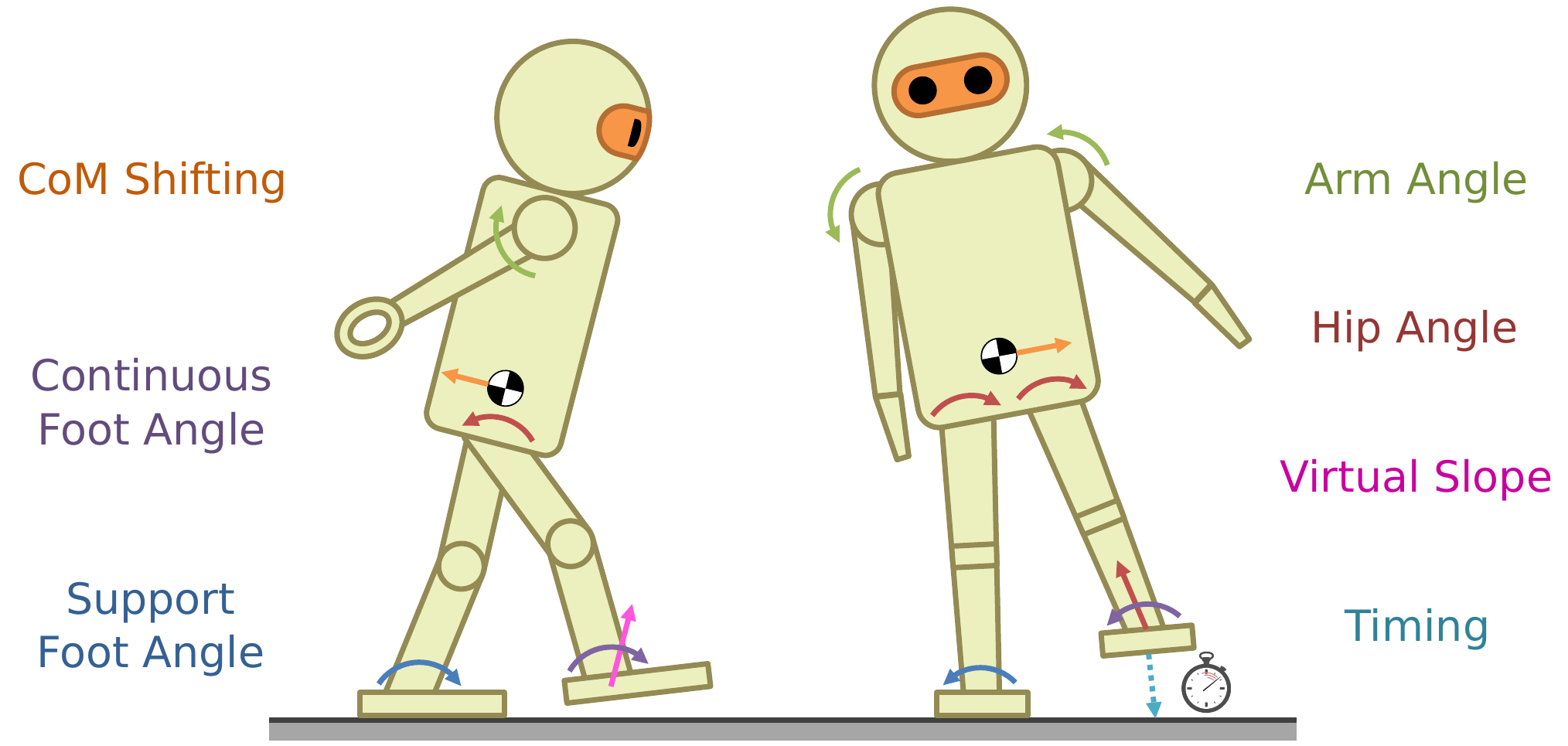}}
\caption{Example fused angle feedback corrective actions, including arm, hip, foot
and foot height components, in both the sagittal (left) and lateral (right) planes.
Step timing is also considered based on the lateral motion.}
\figlabel{basic_feedback}
\vspace{-3ex}
\end{figure}
\section{Reception}

To date, we have built five copies of the \iguhop in our lab, with the parts for 
a further two already printed, and have delivered a complete set of printed 
parts to the University of Newcastle's NUbots RoboCup team. We have demonstrated 
the robots at the RoboCup and various industrial trade fairs, including for 
example the Hannover Messe in Germany and the International Robot Exhibition in 
Tokyo, where the robots had the opportunity to show their interactive side. 
Demonstrations ranged from expressive and engaging looking, waving and idling 
motions, to visitor face tracking and hand shaking. The robots have been 
observed to spark interest and produce emotional responses in the audience.

Despite the recent design and creation of the platform, work groups have already 
taken inspiration from it, or even directly used the open-source hardware or 
software. A good example of this is the Humanoids Engineering \& Intelligent 
Robotics team at Marquette University with their MU-L8 robot. A japanese 
robotics business owner, Tomio Sugiura, has also started printing parts of the 
\iguhop on an FDM-type 3D printer with great success. Naturally, the platform 
also inspired other humanoid soccer teams, like WF Wolves and Baset, to improve 
upon their own robots in certain respects. The \nop, which was a prototype for 
the \iguhop, has also been successfully used in human-robot interaction research 
at the University of Hamburg \cite{barros2014real}.

In 2015, the \iguhop participated for the first time at RoboCup, and was awarded 
the first RoboCup Design Award, based on criteria such as performance, 
simplicity and ease of use. At RoboCup 2016, the platform was an integral part 
of the winning team NimbRo TeenSize (see \figref{robocup_summary}), with a 
combined score of 29:0 over five games. The platform was also awarded the first 
International HARTING Open Source Prize.

\begin{figure}[!t]
\parbox{\linewidth}{\centering
\includegraphics[height=4.0cm]{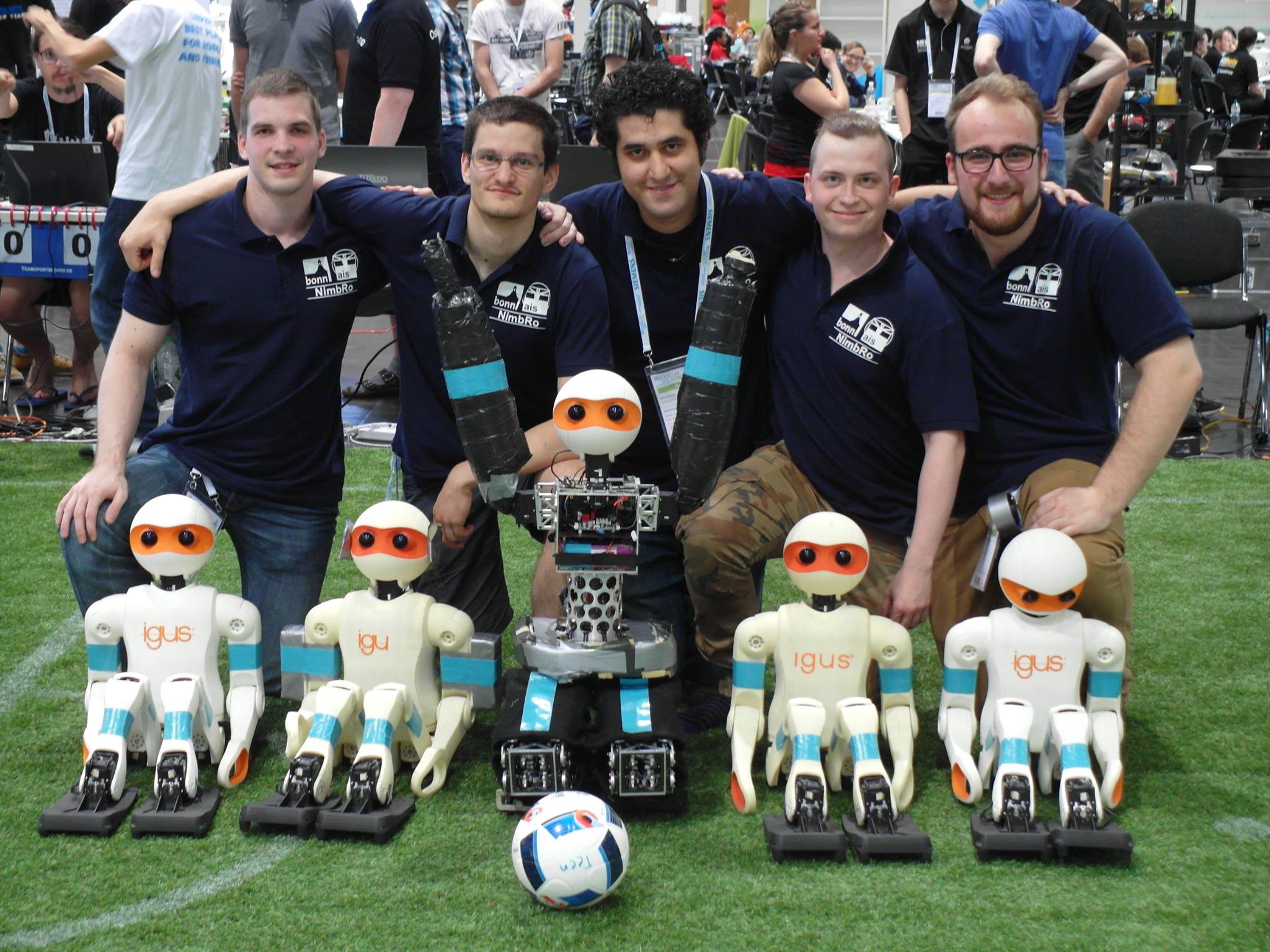}\hspace{0.01\linewidth}
\includegraphics[height=4.0cm]{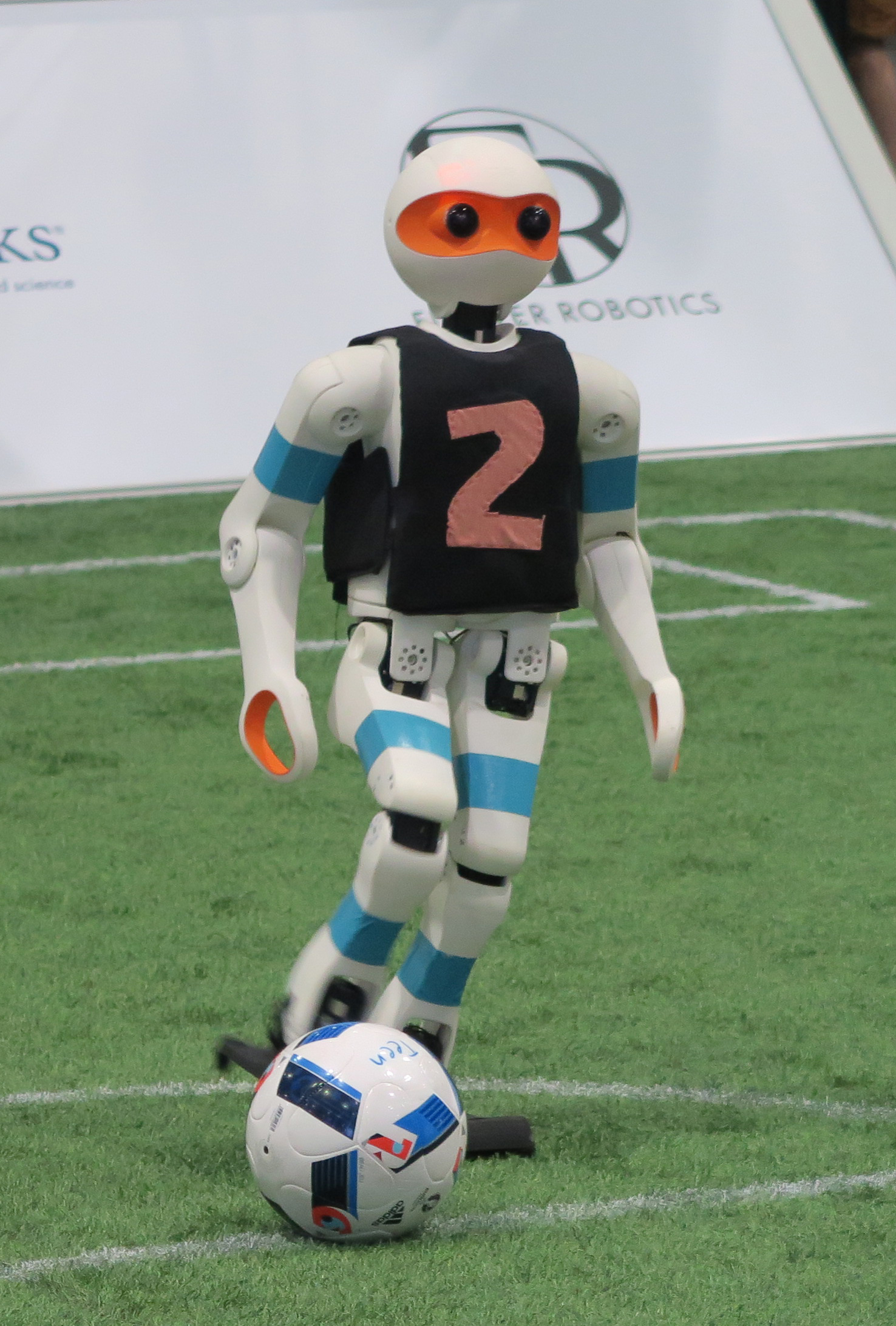}}
\caption{Team NimbRo at RoboCup 2016 in Leipzig.}
\figlabel{robocup_summary}
\vspace{-3ex}
\end{figure}
\section{Conclusions}

The \iguhop represents a significant advancement over its predecessor, towards a 
robust, affordable, versatile and customisable open standard platform. It 
provides users with a rich set of features, while still maintaining modularity 
and flexiblity in the design. The main hardware and software components of the 
robot have been described in this article, and remain a continuous development 
effort. We have released the hardware in the form of print-ready 3D CAD 
files\footnote{\url{https://github.com/igusGmbH/HumanoidOpenPlatform}}, and 
uploaded the software to 
GitHub\footnote{\url{https://github.com/AIS-Bonn/humanoid_op_ros}}. We hope that 
it will benefit other research groups around the world, and encourage them to 
publish their results as a contribution to the open-source community.

\begin{acknowledgements}
We would like to acknowledge the contributions of \igus GmbH, in particular the 
management of Martin Raak towards the robot design and manufacture.
\end{acknowledgements}

\bibliographystyle{spmpsci}
\bibliography{ms}

\end{document}